\documentclass[lettersize,journal]{IEEEtran} 
\usepackage{amsmath,amsfonts}
\usepackage{algorithmic}
\usepackage{algorithm}
\usepackage{array}
\usepackage{textcomp}
\usepackage{stfloats}
\usepackage{url}
\usepackage{verbatim}
\usepackage{graphicx}
\usepackage{cite}
\usepackage{footnote}
\usepackage{comment}
\usepackage{multirow}
\usepackage{amssymb}
\usepackage{subfig}
\usepackage{amsmath}
\usepackage{threeparttable}
\usepackage{balance}
\usepackage{float}
\usepackage{soul}
\usepackage{makecell}
\usepackage{rotating}
\usepackage[table]{xcolor}
\usepackage{fancyhdr}
\begin{document}
\title{Towards Privacy-Preserving Affect Recognition: A Two-Level Deep Learning Architecture}

\author{ Jimiama M. Mase$^{1}$, Natalie Leesakul$^{1}$, Fan Yang$^{3}$, Grazziela P. Figueredo$^{1}$, Mercedes Torres Torres$^{2}$\\
$^1$School of Computer Science, The University of Nottingham, UK\\
$^2$B-Hive Innovations, Lincoln, UK\\
$^3$Hisilicon Research Department, Huawei Technologies Co., Ltd, Shanghai, China\\
}

\maketitle
\begin{abstract} 
Automatically understanding and recognising human affective states using images and computer vision can improve human-computer and human-robot interaction. However, privacy has become an issue of great concern, as the identities of people used to train affective models can be exposed in the process. For instance, malicious individuals could exploit images from users and assume their identities. In addition, affect recognition using images can lead to discriminatory and algorithmic bias, as certain information such as race, gender, and age could be assumed based on facial features. Possible solutions to protect the privacy of users and avoid misuse of their identities are to: (1) extract anonymised facial features, namely action units (AU) from a database of images, discard the images and use AUs for processing and training, and (2) federated learning (FL) i.e. process raw images in users' local machines (local processing) and send the locally trained models to the main processing machine for aggregation (central processing). In this paper, we propose a two-level deep learning architecture for affect recognition that uses AUs in level 1 and FL in level 2 to protect users' identities. The architecture consists of recurrent neural networks to capture the temporal relationships amongst the features and predict valence and arousal affective states. In our experiments, we evaluate the performance of our privacy-preserving architecture using different variations of recurrent neural networks on RECOLA, a comprehensive multimodal affective database. Our results show state-of-the-art performance of $0.426$ for valence and $0.401$ for arousal using the Concordance Correlation Coefficient evaluation metric, demonstrating the feasibility of developing models for affect recognition that are both accurate and ensure privacy.  

% Our results show state-of-the-art performance of $0.4983$ for valence and $0.4232$ for arousal using Concordance Correlation Coefficient evaluation metric, proving that it is possible to develop models for affect recognition that are both accurate and maintain the privacy of the users. 
% Add acknowledgement of Huawei Innovation Research Program (HIRP) and EPSRC
\end{abstract}


\section{Introduction}
Affect recognition is an important domain in artificial intelligence due to its significant health, safety and entertainment potentials~\cite{mase2020evaluating,richman2005positive}. For example, Richman \textit{et al}~\cite{richman2005positive} suggest that negative emotions have a high likelihood of causing chronic stress and cardiovascular diseases, as well as a significant reduction in work performance, while Mase \textit{et al}~\cite{mase2020evaluating} suggest that the emotional states of drivers, both positive and negative, have a significant impact on their driving performance. 

\begin{figure}[htbp]
\centering
\scalebox{0.45}{
\includegraphics{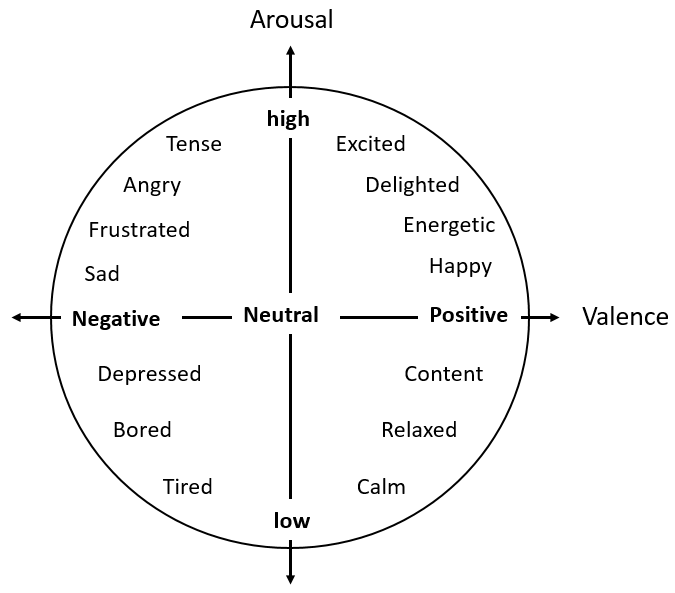}
}
\caption{Two-dimensional description of emotions}
\label{fig:dimensions}
\end{figure}

Images with facial expressions are one of the main data sources for affect recognition in modern studies~\cite{kim2017multi,jain2018hybrid}. Facial expressions are a form of nonverbal communication that helps humans and systems to understand emotional states from facial movements. The main categories of affective states identified by facial expressions are: happy, neutral, sad, surprised, angry, disgusted and anxious~\cite{breuer2017deep}. These affective states are intuitive and simple (i.e. easy to observe and describe), but they fail to represent complex levels of emotions~\cite{sun2014deep}. A more realistic and comprehensive representation of affective states uses a two-dimensional continuous description of emotions (i.e. valence and arousal dimensions)~\cite{ringeval2013introducing} as shown in Fig.~\ref{fig:dimensions}. The valence dimension varies from positive emotions, such as happy and energetic, to negative feelings, such as angry and sad. The arousal dimension ranges from excited to calm moods. 

Deep Learning (DL) methods have been successfully used to detect the affective state of facial expressions~\cite{kim2017multi,jain2018hybrid}. However, the affect recognition systems proposed in previous studies process images in a non-federated manner, which pose serious privacy concerns if the images are accessed by malicious users or organisations in a central database. For instance, a hacker could obtain and assume users' identities~\cite{rathor2013social}, or a malicious organisation may use the imagery demographic information for discriminatory bias~\cite{zerr2012privacy}. In this work, we propose a two-level privacy-preserving strategy. The first level extracts Action Units (AUs) from a database of raw images, discards the images and processes the AUs as AUs safeguard the identities and demographic information of users in case of unauthorised access or misuse of data. The second level employs a Federated Learning (FL) approach where raw images are processed in users' local machines and the locally trained models sent to the main processing machine for aggregation. The main contributions of this study are:

% Furthermore, affective recognition systems need a robust evaluation methodology to ensure accurate and reliable classification systems as misclassification of human emotions may lead to inappropriate interactions and interventions~\cite{jenness2015misclassification}. 

\begin{itemize}
    \item Employing FL in affect recognition using images.
    \item Comparing the prediction, efficiency and privacy performance of non-federated processing of raw images, non-federated processing of anonymised facial features (AUs) and FL of raw images using different variations of Recurrent Neural Networks(RNNs).
\end{itemize}

\section{Background}
\label{background}
In this section, we will first provide an overview of the importance of privacy in affect recognition. Secondly, we will review the literature on processing images and AUs using non-federated deep learning strategies, and present the privacy limitations of the strategies. Later, we will outline all works that explore a FL approach to preserve user privacy and detect emotions.

% , and conclude the section with a description of the RECOLA database, which is the main database used in the literature to detect continuous valence and arousal states.

\subsection{Privacy motivation}
Affective AI technologies are increasingly being adopted and becoming more prevalent in the early 2020s~\cite{mcstay2020emotional}. Such technologies can be deployed in various contexts for different purposes ranging from assistive technology~\cite{bishop2015supporting} to more personalised user experiences, to behaviour manipulation~\cite{poria2017review}. However, in some cases, the use of affective AI can be viewed as causes to privacy-related concerns. For example, the ubiquitous use of facial emotion recognition systems by retailers poses significant privacy challenges~\cite{retailers}. As emotion recognition systems rely on automated facial recognition, retailers are able to identify customers, track their emotions and infer different psychological states, as well as gather and process their personal data, resulting in big datasets for consumer profiling and bias, and in some cases, for price and service discriminations~\cite{wachter2020affinity}. In addition, McStay~\cite{mcstay2016empathic} noted that data derived from someone’s emotional state may be ``intimate'' and sensitive information and can be easily linked back to the person using their image.  

On this account, individual identity should be protected to prevent further exploitation by malicious individuals and companies. 

\subsection{Non-federated deep learning methods}
Different non-federated deep learning architectures have been successfully used to process raw images and predict valence and arousal states~\cite{tzirakis2017end,chao2015long,khorrami2016deep,lee2018spatiotemporal}. For example, Tzirakis \textit{et al}~\cite{tzirakis2017end} reported best valence and arousal recognition performance after training Convolutional Neural Networks (CNNs) coupled with Long Short Term Memory networks (LSTMs) on images from the RECOLA database, while Lee \textit{et al}~\cite{lee2018spatiotemporal} combined features extracted from RECOLA images using 3D CNNs with spatio-temporal features extracted using Convolutional LSTMs to predict valence score. These non-federated CNN approaches require the developer to maintain a database of images for the automatic extraction of non-linear features, and as such, are susceptible to privacy issues if the images are accessed by malicious users or organisations. 

\begin{figure}[htbp]
\centering
\scalebox{0.3}{
\includegraphics{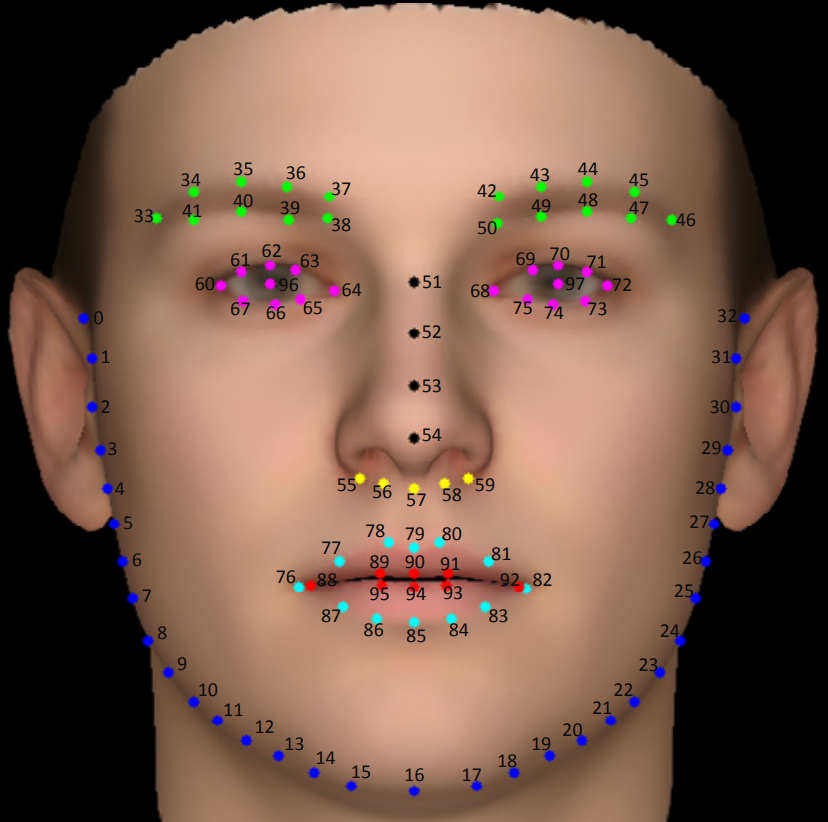}
}
\caption{98 facial landmarks aligned on the face of an avatar depicted from Wu \textit{et al}~\cite{wu2018look}}
\label{fig:faciallandmarks}
\end{figure}

To protect participants' privacy and still accurately identify human emotions, researchers have explored AUs, extracted from facial expressions in images~\cite{chao2015long,ortega2019multimodal,valstar2016avec,han2017strength}. These AUs represent human-observable facial muscle movements, which estimates the intensity of facial movements using facial landmarks. For example, AUs 12 (raising lip corners), 15 (lowering lip corners) and 20 (lip stretch) can be estimated using the facial landmarks on the lips of the avatar face in Fig.~\ref{fig:faciallandmarks}. We also observe from Fig.~\ref{fig:faciallandmarks} that the entire face (thousands of pixels) is reduced to just 98 points containing far less identifiable information which are further processed to obtain the AUs. In typical non-federated AU approaches, developers extract AUs from the images, create a database of AUs, discard the images and use the AU database to train their models. Such methods~\cite{chao2015long,ortega2019multimodal,valstar2016avec,han2017strength} have shown to be more privacy-preserving but less accurate compared to processing the images. In addition, AUs are not completely private as a recent study by Fan \textit{et al}~\cite{fan2021demographic} demonstrates relationships between AUs and demographic factors such as race, gender and age.

\subsection{Federated deep learning methods}
In order to provide complete user privacy and avoid misuse of their identities, one possible solution is to employ a FL approach~\cite{yang2019federated}. FL is a technique that allows a ML model to be trained without collecting data. This is done by the collaborative training of multiple ML models on users local machines (local models) where their personal data resides and sending the trained models (i.e model weights) back to the developer's machine (central model) for aggregation. Different ensemble methods can be employed to aggregate the model weights depending on the problem such as mean, median, and weighted average. The central model updates its weights using the aggregated weights and sends the updated weights back to the local models. This process keeps the local training data private and confidential. FL methods have been employed in speech emotion recognition~\cite{latif2020federated} and detection of depression using mobile health data~\cite{xu2021fedmood} but not in Facial Emotion Recognition (FER). To the best of our knowledge, the only study that mentions FL for FER is Chhikara \textit{et al}~\cite{chhikara2020federated}. However, the study does not implement FL nor present any FL results. They simply mention FL as a privacy solution to their multi-modal affect recognition approach.

In this paper, we implement a two-level privacy preserving strategy consisting of non-federated AUs and FL of images using deep neural networks. We assess their effectiveness in terms of accuracy, efficiency and privacy by comparing their performance with a non-federated image processing strategy. We test the approaches on RECOLA, a comprehensive multimodal affect database with continuous valence and arousal states.

% amongst facial AUs extracted from the raw images. Research in affective computing is still very limited as people are unwilling to take part in studies due to privacy concerns. There are currently several publicly available facial AU detectors~\cite{magdin2018real} that are capable of automatically extracting numerous AUs for analysis. These detectors can be used to preserve users' privacy and advance research in emotion recognition. 

\section{Methodology}
\label{Methodology}
In this section, we introduce the privacy-preserving schemes proposed, including (1) non-federated processing of extracted facial features (AUs), and (2) FL using the raw images. The processing modules for those schemes use RNNs to learn temporal relationships between facial features and emotional states~\cite{jain2018hybrid,zhu2017dependency}. 

\subsection{Level 1: Non-federated processing of action units}
\begin{figure*}[htbp]
\centering
\scalebox{0.35}{
\includegraphics{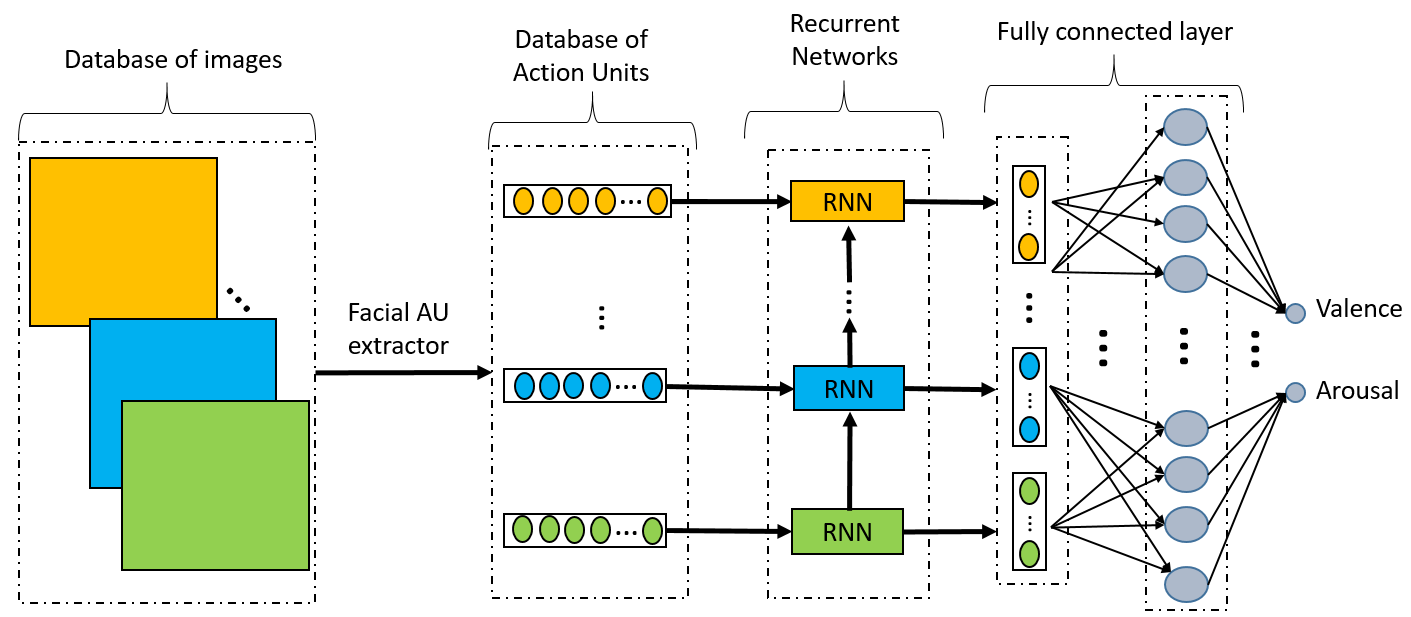}
}
\caption{A non-federated deep learning strategy for affect recognition using action units}
\label{fig:method_AUs}
\end{figure*}

Fig.~\ref{fig:method_AUs} shows the first level of our privacy-preserving architecture that processes AUs in a centralised manner. AUs are extracted from a database of images using facial landmark detectors (e.g. OpenFace AU~\cite{baltrusaitis2018openface}) and the images are discarded to protect users' identities and demographic information as the AUs are free of human faces and demographic information. Later, the AUs are pre-processed by transforming them to similar scales for better performance (e.g. normalisation). The pre-processed AUs serve as the input to the processing module. We utilise RNNs in the processing module to capture the temporal relationships among the sequential AUs and feed the temporal features to fully-connected neurons. The fully-connected neurons learn the non-linear relationships between the temporal features and the continuous affective states. 

\begin{figure*}[htbp]
\centering
\scalebox{0.45}{
\includegraphics{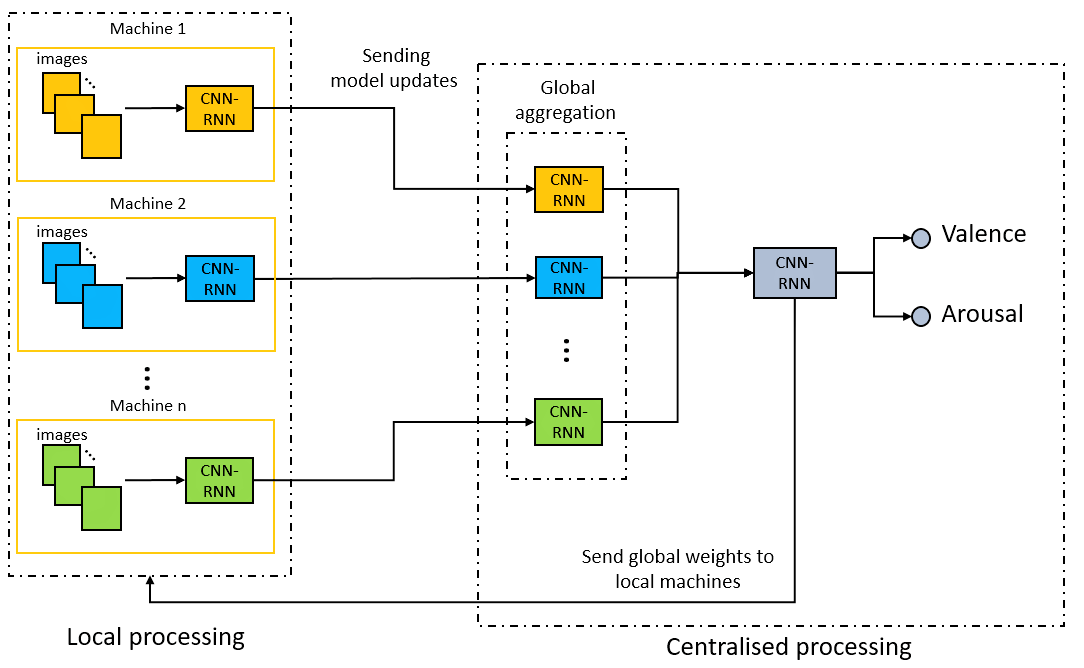}
}
\caption{A federated learning approach for affect recognition using images}
\label{fig:method_FL}
\end{figure*}

\subsection{Level 2: Federated learning using images}
Level 2 of our architecture uses a FL approach to process users' images at their local machines and their trained models sent to the central processing module for aggregation as shown in Fig.~\ref{fig:method_FL}. The local and central processing machines should have the same model implementation to enable easy aggregation of the trained models. For simplicity, we adopt a mean aggregation strategy where the weights of the trained models are averaged to represent the aggregated weights (global weights). It is important to note that different aggregation strategies can be explored to merge the weights e.g. the central processing module could maintain \textit{n} global weights for the \textit{n} local machines where the different global weights are weighted averages of the local machines. We implement CNNs coupled with RNNs to process the images at the local machines. The CNNs and RNNs are trained together end-to-end. 

Training occurs simultaneously across the machines. After each training iteration, the locally trained models (i.e. model weights) are sent to the central processing module for aggregation. The centrally aggregated weights  are sent back to the local machines to update their weights for the next training iteration. The local training, central aggregation and local weight update processes are repeated until the training process is completed. 

\section{Experimental design}
\label{experidesign}
This section first describes the RECOLA database and presents the deep learning methods selected for the processing modules of our architecture along with their hyper-parameter configurations. Furthermore, it defines the Concordance Correlation Coefficient (CCC) metric used for evaluating the accuracy of the models and concludes with the evaluation protocols for the experiments. 

\subsection{RECOLA database}
\label{database}
The RECOLA (Remote COLlaborative and Affective interactions) database~\cite{ringeval2013introducing} is the most popular and comprehensive affective dataset with continuous response variables (i.e. valence and arousal). The database consists of images, AUs, audio, ECG and EDA datasets for 23 participants. The data was collected during spontaneous and naturalistic interactions between the participants when performing collaborative tasks. The database also contains the ground truth continuous labels (valence and arousal) that range from -1 to +1 with a step size of 0.01. The annotations were carried out by six annotators. In this study, we will explore the image and AU datasets of RECOLA. The AU dataset consists of 40 AUs.

\subsection{Model selection and hyper-parameter configuration}
\begin{figure*}[htbp]
\centering
\scalebox{0.35}{
\includegraphics{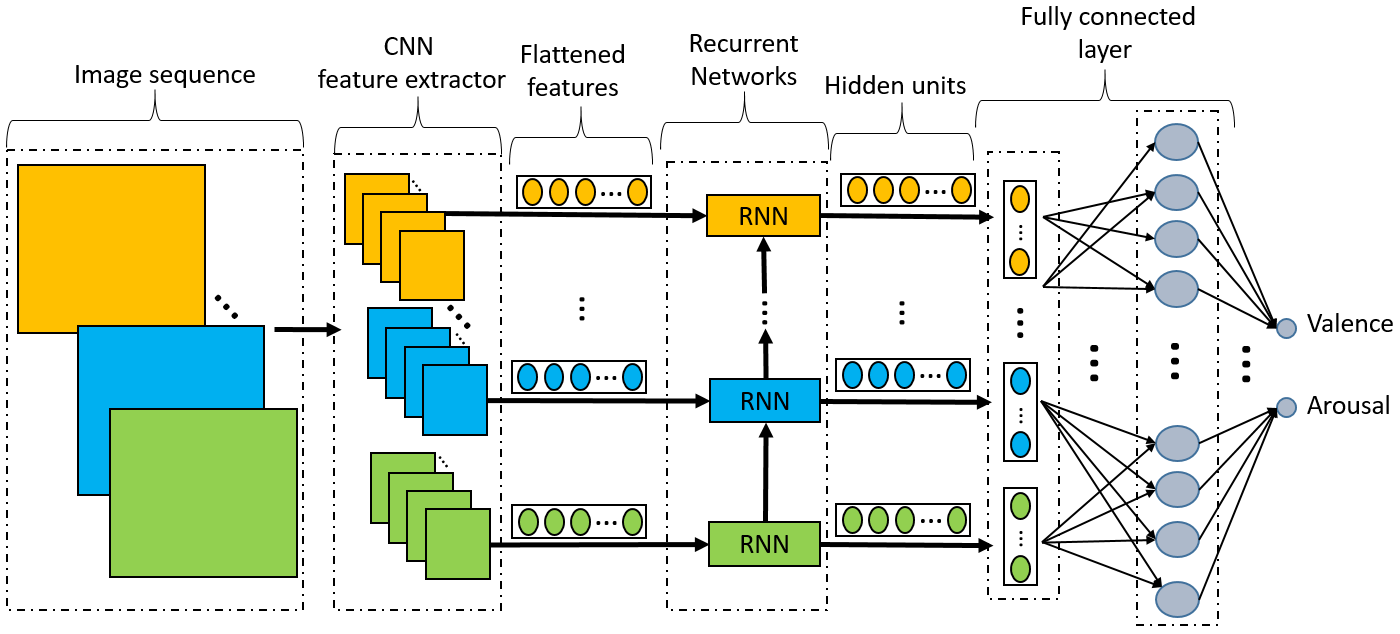}
}
\caption{A deep learning approach for processing images}
\label{fig:c-slstm}
\end{figure*}

We explore three state-of-the-art RNN models to detect valence and arousal: simple RNNs~\cite{jain2018hybrid}, Bi-directional Gated Recurrent Units (BiGRUs)~\cite{chung2015gated}, and Bi-directional Long Short Term Memory networks (BiLSTMs)~\cite{tzirakis2021end}. We choose these models due to their remarkable performance in time series or sequential analysis~\cite{mase2020benchmarking}. To process the raw images in level 2 of our architecture, we employ shallow residual convolutional networks (i.e. ResNet18)~\cite{bengio1994learning} due to their remarkable training efficiency (fewer number of layers compared to other state-of-the-art CNNs) and prediction performance~\cite{mase2020benchmarking}. The ResNets are pre-trained on the ImageNet dataset~\cite{russakovsky2015imagenet} to take advantage of its large size (transfer learning). Later, we remove the fully connected layers of the networks and use their output feature maps as inputs to the RNN networks. The performance of our privacy-preserving strategies are compared with the non-federated processing of raw images, where a database of user images is maintained. For this, we utilise a deep learning method similar to that employed for processing the images in FL i.e. ResNets coupled with RNNs, trained together end-to-end as shown in Fig.~\ref{fig:c-slstm}.

When training the networks, we minimise the Mean Squared Error (MSE) between the predicted affective states and annotated affective states, and we use Adam Stochastic Gradient Descent to optimise the loss function (MSE), which is a fast optimisation algorithm for deep neural networks. The RNN networks consist of the following hyper-parameters: learning rate, hidden layers, sequence length, number of recurrent layers and fully connected layers. The learning rate controls how the weights are updated with respect to the estimated error. If the learning rate is very low, the learning process will be slow as the updates will be very small, and if the learning rate is very high, the weight updates will be very large which can lead to divergence. We train the models using popular learning rates, 0.001, 0.0001, and 0.00001. The hidden size represents the number of hidden units within each recurrent memory cell. We explored 8, 16, 64, 128, 256 and 512 hidden sizes. We also explored 50, 100, 200, 400, 600, 800, 1000 and 2000 AU sequence lengths, and 4, 8, 16, 32 image sequence lengths. The following number of recurrent memory cells (recurrent layers) were evaluated: 1, 2, 4, 6 and 8. Lastly, one fully connected layer was used in the networks consisting of 10 neurons with 2 output neurons for valence and arousal affective states. Table~\ref{tab:config} presents the optimal hyper-parameter configurations of the architectures after evaluating the validation loss using the above selected hyper-parameters.

\begin{table*}[hbt!]
  \centering
  \caption{Hyper-parameter configuration of models}
  \scalebox{0.9}{
    \begin{tabular}{c|c|c|c|c|c}
     \textbf{Method} & \textbf{Networks} & \textbf{Learning rate} &  \textbf{Sequence length}& \textbf{Hidden size} & \textbf{Number of layers} \\
     \hline
    \cline{2-6} & BiGRU  &	0.0001 & 16 & 8 & 1 \\
    \cline{2-6}
     Raw Images & BiLSTM  & 0.0001 & 16 & 8 & 1 \\
    \cline{2-6}
     & RNN   &  0.0001 & 16 & 8 & 1\\
    \hline
        \cline{2-6} & BiGRU  &	 0.0001 & 600   & 512   & 6 \\
    \cline{2-6}
     Action Units & BiLSTM  & 0.0001 & 600   & 128   & 6 \\
    \cline{2-6}
     & RNN   & 0.0001 & 2000  & 128   & 2  \\
    \hline
    \cline{2-6} & BiGRU  & 0.0001 & 8 & 128  & 6  \\
    \cline{2-6}
     Federated Learning  & BiLSTM & 0.0001  & 8 & 128  & 6  \\
    \cline{2-6}
     & RNN   & 0.0001  & 8 & 128  & 6\\
    \hline
    
    \end{tabular}}%
  \label{tab:config}%
\end{table*}%

\subsection{Evaluation metrics}
\label{evalmetrics}
\begin{figure}[!htbp]
\includegraphics[width=\linewidth]{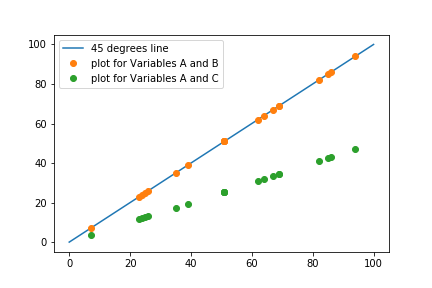}
      \caption{An example to compare Pearson’s correlation and CCC}
      \label{fig:ccc}
\end{figure}

For performance evaluation, we use Concordance Correlation Coefficient (CCC). CCC is the correlation between two variables that fall on the 45 degrees line through the origin. Similarly to Pearson's correlation coefficient, CCC measures how closely related two variables are in linear fashion, but it also calculates the degree of correspondence (agreement) between the two variables by measuring their fitness to the line passing through the origin with a slope of 1. It is said to be more robust than Pearson's correlation as it measures both covariation and correspondence. Fig.~\ref{fig:ccc} shows two plots (orange and green) with Pearson's correlation coefficients of 1 but the orange plot has a CCC of 1 while the green plot has a CCC of 0.403 due to its disagreement with the 45 degree line. CCC ranges from -1 to 1, with perfect concordance at 1 and perfect discordance at -1.

CCC is calculated as follows:

\begin{align*}
CCC = 2\rho\sigma_x\sigma_y/\sigma_x^2 + \sigma_y^2 + (\mu_x - \mu_y)^2\\
\end{align*}

where $\mu_x$ and $\mu_y$ are the means for the two variables and $\sigma_x^2$ and $\sigma_y^2$ are their corresponding variances. $\rho$  is the correlation coefficient between the two variables.

\subsection{Evaluation protocol}
First, the ground truth valence and arousal values are obtained by averaging the annotations from the six annotators. Secondly, we employ \textit{k}-fold cross validation to evaluate the models. The dataset is split by participants to prevent overfitting. In our experiments, we select \textit{k} = 8 i.e. data split into 8 folds with each fold consisting of data for 2-3 participants depending on the split. The training process is repeated \textit{k} times to produce \textit{k} trained models and during each training process, one fold is left out for evaluating the model and the remaining folds are used to train the model. The average CCC amongst the k evaluated models gives the overall performance of the method across the entire dataset. The higher the value of k the more computationally expensive is the training process, however, the more robust and accurate is the model's performance. For a more realistic implementation of FL, we use each participant as a local machine and divide the total training time by the number of participants to represent synchronous local processing. For example. using 8-fold cross validation on 23 participants, we have data for 20 or 21 participants (local machines) for training and data for the remaining 2 or 3 participants kept aside for evaluating the global model. All experiments are executed on a Graphics Processing Unit (GPU) using 4 CPU cores and 6GB RAM. Our code is implemented in Pytorch with an epoch size of 100 for each experiment.

\begin{table*}[hbt!]
  \centering
    \caption{Average CCC for predicting valence and arousal using variations of RNN models on RECOLA datasets (best performance in bold).}
  \scalebox{0.9}{
    \begin{tabular}{c|c|c|c}
     \textbf{Affect} & \textbf{Deep} &  \textbf{Average of} & \textbf{Average of} \\
       \textbf{Recognition Method} &\textbf{Neural Network}  & \textbf{Valence CCC} & \textbf{Arousal CCC} \\
     \hline
    \cline{2-4} & CNN-BiGRU  &		0.415  &	0.504\\
    \cline{2-4}
    Non-federated Processing & CNN-BiLSTM  & \textbf{0.476} & \textbf{0.515}\\
    \cline{2-4}
    of Raw Images & CNN-RNN  & 	0.471 &	0.486 \\
    \hline
        \cline{2-4} & BiGRU & 0.347 &	0.401 \\
    \cline{2-4}
    Non-federated Processing & BiLSTM  & 0.269	& 0.365\\
    \cline{2-4}
    of Action Units & RNN   & 0.183 &	0.228\\
    \hline
    \cline{2-4} & CNN-BiGRU  &	0.393 &	0.273 \\
    \cline{2-4}
    Federated Processing& CNN-BiLSTM  & 0.426 &	0.390 \\
    \cline{2-4}
    of Raw Images & CNN-RNN   & 0.304 &	0.197 \\
    \hline
    
    \end{tabular}}%
  \label{tab:model_comparison}%
\end{table*}%

\begin{table*}[hbt!]
  \centering
  \caption{Model training time, inference time, and size for the best performance RNNs (best performance in bold).}
  \scalebox{0.9}{
    \begin{tabular}{c|c|c|c|c|c|c|c|}
     \textbf{Affect recognition} & \textbf{Data Privacy} & \textbf{Average of} & \textbf{Average of} & \textbf{Training time} &  \textbf{Model size} &\textbf{Inference time using} &  \textbf{Inference time using}\\
      \textbf{method} & \textbf{Level} & \textbf{Valence CCC} & \textbf{Arousal CCC} & \textbf{(mins)} & \textbf{(MB)}&  \textbf{100 images(secs)}  &\textbf{for 500 images(secs)}\\
       \hline
    Raw Images  & Low & \textbf{0.476} & \textbf{0.516} & 315.4 & \textbf{43} & 17.509 & 22.921\\
     \hline
    Action Units  & Average & 0.347 &	0.401 & \textbf{141.5} & 98 & \textbf{1.351} & \textbf{2.962}\\
     \hline
    Federated Learning  & \textbf{High} & 0.426 & 0.390 & 599.2 & 53 & 19.720 & 26.860 \\
    \end{tabular}}%
  \label{model_efficiency}%
\end{table*}%

\begin{table*}[hbt!]
  \centering
    \caption{Comparison of valence and arousal predictions between our proposed methods and other studies using RECOLA datasets (best performance in bold).}
  \scalebox{0.9}{
    \begin{tabular}{c|c|c|c}
     \textbf{Affect} & \textbf{Type of Machine} &  \textbf{} & \textbf{} \\
       \textbf{Recognition Method} &\textbf{Learning Model}  & \textbf{Valence CCC} & \textbf{Arousal CCC} \\
     \hline
    \cline{2-4} & CNN + LSTM~\cite{tzirakis2017end}  &	\textbf{0.620} & 0.435 	\\
        \cline{2-4} & DNN~\cite{ortega2019multimodal}  &	0.379  & 0.464 	\\
    \cline{2-4}
     Non-federated Processing & CNN + LSTM~\cite{chao2015long}  & 0.538 & 0.336  	\\
    \cline{2-4}
    of Raw Images & CNN + RNN~\cite{khorrami2016deep}  & 0.474 & - 	\\
     \cline{2-4} & 2D CNN + ConvLSTM + 3D CNN~\cite{lee2018spatiotemporal} & 0.546 & -\\

    \cline{2-4} & \textbf{Our CNN + BiLSTM} & 0.476 & \textbf{0.514} \\
    \hline
    \cline{2-4} & LSTM~\cite{chao2015long} & 0.483 &	0.137 \\
    \cline{2-4}
    Non-federated Processing & SVM~\cite{valstar2016avec}  & \textbf{0.507} & 0.272\\
    \cline{2-4}
    of Action Units or Landmarks & BiLSTM + SVM~\cite{han2017strength} & 0.394 &	0.265\\
    \cline{2-4} & \textbf{Our BiGRU} & 0.347 &	\textbf{0.401} \\
    \hline
    \cline{2-4} &  &  &	 \\

    Federated Processing & \textbf{Our CNN + BiLSTM} & \textbf{0.426} &	\textbf{0.390}\\

    of Raw Images &   &  &	 \\
    \hline
    
    \end{tabular}}
    \begin{tablenotes}
      \centering
      \item Note: A dash is inserted if the results were not reported in the original papers.
    \end{tablenotes}%
  \label{tab:results}%
\end{table*}%

\section{Results and Discussion}
\label{results and discussion}

\subsection{Comparison of proposed methods}

We implemented three state-of-the-art RNN models (i.e., RNN, BiGRU, and BiLSTM)  for each method and evaluated their performance using CCC coupled with cross-validation on the RECOLA image and AU datasets. Table~\ref{tab:model_comparison} shows the average CCC for valence and arousal after evaluating the models using the best hyper-parameters shown in Table~\ref{tab:config}. The bold values represent the best model performance for valence and arousal affective states. Overall, the non-federated processing of raw images shows best valence and arousal predictions, followed by the federated processing of raw images. These strategies that process raw images outperform the processing of AUs due to the loss of spatial information in the AUs. CNNs coupled with BiLSTMs show best performance for non-federated processing of images with \textit{0.476} average CCC for valence and \textit{0.515} for arousal. Next, the processing of AUs shows similar arousal prediction performance compared to the federated processing of images. In addition, we observe that LSTMs outperform GRUs when processing the raw images similar to results from other studies that analyse raw images~\cite{mase2020benchmarking}. However, for AU processing, GRUs show better performance compared to LSTMs. This is due to the efficiency of GRUs in processing smaller datasets or feature sets compared to LSTMs as only 40 AUs are extracted by the facial landmark extractor while 512 features are extracted by the convolutional networks. 

Table~\ref{model_efficiency} presents the efficiency results of the best performing models in terms of training time, inference time and model size. We observe that processing AUs has the least training and inference times due to a smaller feature set (which reduces the complexity of the network) and lack of the convolutional feature extraction layer. This makes the AU processing modules more suitable for real-time affect recognition systems such as, real-time monitoring of patients' affective states for early intervention and assistance~\cite{song2020spectral}. However, the predictive accuracy of processing AUs is lower compared to the non-federated processing of raw images for both valence and arousal. The non-federated processing of raw images shows better accuracy in predicting valence and arousal compared to AUs and FL at the detriment of users' privacy and inference time. FL best preserves users' privacy and identity compared to the other methods as data is maintained in users' local machines, however, its training time is significantly higher, which can further increase if the processing at the local machines is not done synchronously. Lastly, FL's CCC results are inferior to the non-federated processing of images due to limited data at the local machines.

\subsection{Comparison with other studies}

In Table~\ref{tab:results}, we compare the performance of our models with other studies that employ machine learning methods on the RECOLA image and AU datasets for affect recognition. For the non-federated processing of raw images, we observe that~\cite{tzirakis2017end,ortega2019multimodal,chao2015long,lee2018spatiotemporal} show better valence recognition results compared to our model, with Tzirakis \textit{et al}~\cite{tzirakis2017end} having the best CCC valence (\textit{0.620}). However, our model shows best arousal accuracy with a CCC value of \textit{0.514}. Those studies also explored different architectures of CNNs coupled with LSTMs, however, they are limited in their model evaluation strategy (i.e. train-test split) that prevents a comprehensive exploration of the data and may lead to less accurate or biased results. 

Furthermore, the processing of AUs and facial landmarks by previous studies~\cite{chao2015long,valstar2016avec,han2017strength} show better CCC results in predicting the valence dimension. Valstar \textit{et al}~\cite{valstar2016avec} presented best valence CCC results of \textit{0.507} using support vector machines. Our model shows a contrary performance as our arousal prediction results are better than valence and outperforms the arousal accuracy of the other studies (\textit{0.401}). This is due to the remarkable performance of GRUs in processing small feature sets. Next, storing the anonymised AUs is more secured in terms of privacy compared to raw images. Therefore, in order to maintain a database of images with facial expressions, requires appropriate security levels and systems to safeguard the data, which can be challenging to implement. The trade-off between efficiency and privacy is at the cross road. Consequently, from a privacy-compliant perspective, it could be argued that storing raw images may not be necessary if other alternative methods are available and extracting AUs could be considered as a data anonymisation technique for images to protect the identities and demographic information of users, and an alternative method for affect recognition. 

Lastly, we could not find any study in the literature that explores FL to process raw images for FER. As a result, we show the performance of our FL architecture that uses CNNs coupled with BiLSTMs as a benchmark for future research on FL and privacy-preserving deep learning techniques for FER. By taking into account the privacy benefits of FL as well as their promising results (i.e. valence = 0.426 and arousal= 0.390), further research is required to advance FL for affect recognition.

\section{Conclusion}
\label{Conclustion and Future Works}
In this paper, we prioritised privacy in facial emotion recognition by presenting a two-level privacy-preserving architecture consisting of: (1) a deep learning model for processing anonymised facial features (Action Units) from facial expressions to preserve users' identities, and (2) a federated deep learning approach that aggregates locally trained models on raw images. We implemented three variations of RNNs and compared the models’ performance including the non-federated processing of images on the RECOLA databases.  Our results show state-of-the-art performance of $0.426$ for valence and $0.401$ for arousal using Concordance Correlation Coefficient evaluation metric using the privacy-preserving architecture. 

For future work, we plan to improve the performance of these models by combining and fusing data from other modalities while still maintaining the privacy proactive nature of our system as well as promoting responsible technology as ``data protection by design and by default''. For example, extracting and combining acoustic features with the AUs or a federated learning approach to aggregate locally trained models on audio-video data. We also intend to optimise the models, and explore spiking neural networks to reduce model complexity for efficient use in real-time emotion detection systems.

% \section*{Acknowledgement}
% This work was supported by the Horizon Centre for Doctoral Training at the University of Nottingham (UKRI Grant No. EP/L015463/1), the Engineering and Physical Sciences Research Council (DigiTOP; EP/R032718/1), and Huawei Technologies Co., Ltd.

\bibliographystyle{plain}
\balance{{\bibliography{Privacy-Preserving.bib}}

\end{document}

